\documentclass[runningheads]{llncs}
\usepackage[T1]{fontenc}
\usepackage{soul}
\usepackage{booktabs} 
\usepackage[table]{xcolor}
\usepackage{graphicx}
\usepackage{hyperref}
\usepackage{color}

\newcommand\sigWorse[1]{\underline{\textit{#1}}}

\begin{document}
\title{Generational Computation Reduction in Informal Counterexample-Driven Genetic Programming}
%
%
\author{Thomas Helmuth\inst{1}\orcidID{0000-0002-2330-6809} \and
Edward Pantridge\inst{2}\orcidID{0000-0003-0535-5268} \and
James Gunder Frazier\inst{1}\orcidID{0009-0008-9549-1485} \and
Lee Spector\inst{3,4}\orcidID{0000-0001-5299-4797}
}
%
%
\institute{Hamilton College, Clinton NY 13323, USA \email{\{thelmuth,jgfrazie\}@hamilton.edu} \and
Real Chemistry, Boston MA 02111, USA \email{ed@swoop.com} \and
Amherst College, Amherst MA 01002, USA \email{lspector@amherst.edu} \and
University of Massachusetts, Amherst MA 01003, USA}

%
\maketitle              
\begin{abstract}
Counterexample-driven genetic programming (CDGP) uses specifications provided as formal constraints to generate the training cases used to evaluate evolving programs. It has also been extended to combine formal constraints and user-provided training data to solve symbolic regression problems. Here we show how the ideas underlying CDGP can also be applied using only user-provided training data, without formal specifications. We demonstrate the application of this method, called ``informal CDGP,'' to software synthesis problems. Our results show that informal CDGP finds solutions faster (i.e. with fewer program executions) than standard GP. Additionally, we propose two new variants to informal CDGP, and find that one produces significantly more successful runs on about half of the tested problems. Finally, we study whether the addition of counterexample training cases to the training set is useful by comparing informal CDGP to using a static subsample of the training set, and find that the addition of counterexamples significantly improves performance.

\end{abstract}

\section{Introduction}

The bulk of the computational effort required for genetic programming (GP) is expended in the evaluation of programs in the evolving population. Typically, each program is evaluated on many inputs, which are generally referred to as ``fitness cases'' or ``training cases.'' In most prior work, 
all available cases are used to evaluate each program.

Two recent developments in GP have offered new approaches to handling training cases that appear to provide significant advantages. One of these methods uses only a small, random sub-sample of the available cases each generation. This ``down-sampling'' saves significant computational effort per program evaluation, allowing one to run the evolutionary system for more generations with the same computational budget, leading to significant improvements in problem-solving power~\cite{Hernandez:2019:GECCOcomp,Ofria:2019:GPTP,Helmuth2020explaining}.

A second method, counterexample-driven genetic programming (CDGP), generates training cases using formal specifications that must be provided for the problem to be solved~\cite{Krawiec:2017:GECCOa,Bladek:EC,Krawiec:2018:IJCA,Bladek:2019:GECCO,Bladek:IEEETEC:2023,Welsch:GECCO:2020}. In particular, it is able to generate training cases that are not correctly solved by the evolving programs, adding these cases to a growing training set. These ``counterexamples'' provide more focused guidance to the evolutionary process than do random test cases, and appear to direct evolution more specifically to master aspects of the target problem that are not properly handled by individuals in the current population. 
While CDGP has been applied to constrained problem domains where it is possible to check whether any given program satisfies the given formal specifications, it is impossible to check whether programs over a Turing-complete language satisfy given formal specifications~\cite{rice:1953,hopcroft:ullman:1979}. Therefore, CDGP cannot be applied directly to general program synthesis problems, where GP evolves programs that may include looping or recursion and access to potentially unbounded storage.

In this paper, we describe a novel method that builds on ideas of down-sampling the training data and CDGP by extending the idea of counterexamples to not require formal specifications.\footnote{This paper expands on a poster paper that we published in GECCO 2020~\cite{Helmuth:2020:GECCOcompa}.} The approach that we describe, ``informal CDGP'' (iCDGP), evaluates individuals during evolution using only a small sub-sample of the user-provided training cases, like down-sampled GP, allowing more individuals to be assessed within the same computational budget. When training cases are added to the training set, they are not chosen randomly,  but rather are chosen to be counterexamples for the best individuals in the current population. This allows iCDGP to direct evolution in much the same way as CDGP, but without requiring that the user provide formal specifications for solutions to the target problem.

We test iCDGP on a set of general program synthesis benchmark problems, which require evolving programs in a Turing-complete language~\cite{Helmuth:2015:GECCO}.
Initial experiments found that many times GP was not able to find a program that passed all training cases, meaning no new cases were added.
We develop two new variants of iCDGP, and find that one in particular outperforms standard GP; this variant ensures that new cases are added to the training set throughout evolution, whether or not a program is found that passes the current training set. The second variant limits the size of the training set, motivated by making better use of the computational budget; this variant does not show as much empirical promise.


\section{Related Work}

This work takes its motivation from and builds on counterexample-driven GP and down-sampled lexicase selection. We describe each of those techniques in detail, and then discuss other related work.

\subsection{Counterexample-Driven GP}

CDGP uses specifications provided as formal constraints in order to generate the training cases used to evaluate a population of evolving programs~\cite{Krawiec:2017:GECCOa,Bladek:EC,Krawiec:2018:IJCA}. CDGP was extended to use both formal constraints and user-provided training data to solve symbolic regression problems~\cite{Bladek:2019:GECCO,Bladek:IEEETEC:2023}. Additionally, CDGP has been combined with synthesis through unification, which allows it to partially decompose parts of problems into subproblems to solve~\cite{Welsch:GECCO:2020}.

CDGP evaluates individuals in the population against both a set of automatically generated training cases and the provided formal constraints. The training set is the primary method of evaluating individuals for parent selection, and the formal constraints are used to generate new training cases when necessary. 
The CDGP algorithm proceeds as follows:
The training set is empty at the start of evolution. Then, each generation, every individual is evaluated on each training case in the training set. If any individual passes all of the training cases\footnote{Every individual in the first generation passes the training set, since it is initially empty.}, CDGP uses a Satisfiability Modulo Theories (SMT) solver to test the individual on the problem's formal constraints. If the program passes the formal constraints, evolution stops because it has found a solution. If the individual fails a constraint, the SMT solver returns a counterexample in the form of a new case that the program does not pass. CDGP adds this case to the training set for the next generation. This process continues until it either finds a solution or reaches a maximum number of generations.

In standard CDGP, a new counterexample case is added to the training set only when a program passes all current training cases. However, one extension of CDGP adds a fitness threshold $q \in [0, 1]$ that specifies the proportion of the training cases that an individual must pass before running the SMT solver on it to produce a new counterexample case~\cite{Bladek:EC}. This allows new cases to be added earlier, giving more search gradient for evolution to follow without having to find a program that passes all cases in the current training set. A value of $q = 1.0$ is equivalent to the standard CDGP, since it adds a new case only when an individual passes all current cases. 
On the basis of experimentation, the developers of this technique recommend a value for $q$ in the range $[0.75, 1]$.

We are unaware of any previous work employing counterexamples without the use of formal specifications, as we do here.

\subsection{Down-sampled lexicase selection}

While down-sampling of training data has been occasionally used in GP, it has recently been studied in the program synthesis domain when using lexicase parent selection~\cite{Hernandez:2019:GECCOcomp,Ofria:2019:GPTP,Helmuth2020explaining,Helmuth:2022:ALife}. In this setting, the training set is down-sampled to include a random subset of the training cases each generation. Down-sampled lexicase selection reduces the cost to evaluate each individual, with the same motivation as iCDGP.

Multiple studies have examined why and where down-sampled lexicase performs well~\cite{Helmuth:2022:ALife,Schweim:2022:CEC,Hernandez2022}. In evolutionary robotics, down-sampled lexicase selection has been used to limit the costs of robotics simulations~\cite{moore:2017:ecal,moore:2018:alife}. More recently, work has been done to make informed decisions when selecting the cases that appear in the subsampled training set \cite{Boldi:2023:GECCOcompanion,boldi2023problem,boldi_informed_2023}.

\subsection{Other Related Work}

Guiding learning with counterexamples that modify a training set has been recently explored in machine learning~\cite{dreossi:2018,sivaraman:2020}. Implementations of this technique require an error table to be constructed from the model's misclassified data points from training, which is then used as specifications for how to construct counterexamples to train the model. 

Metamorphic testing has been applied to GP to extend the usefulness of each case in exposing undesirable behaviors in candidate solutions without needing to include more cases to train on~\cite{sobania:2023}. To apply metamorphic testing to a problem however, a user must first identify a metamorphic relation a solution program's output must exhibit; these metamorphic relationships are related to but different from the formal specifications required by CDGP.

The core of iCDGP has been used to develop human-driven genetic programming for program synthesis, in which a user is responsible for providing the initial training set and for verifying whether or not generated cases are counterexamples to a potential solution. A prototype system has shown promise on some basic program synthesis problems~\cite{Helmuth:2023:GECCOcompanion}.

\section{Informal Counterexample-Driven GP}

Informal counterexample-driven GP (iCDGP) borrows motivation from CDGP, but deviates in some significant and novel ways. Specifically, we aim to adapt the core concept of a small training set that grows with added counterexamples. 
Since we do not have formal specifications, we instead expect the problem to be defined by a full training set of input/output examples, typically numbering 100 or more, which we call $T$.

In iCDGP, we use an active training set, $T_A \subseteq T$, that GP uses to evaluate the individuals in the population. In all of our experiments, $T_A$ initially contains 10 random training cases from $T$, although other sizes could be used. During evolution, if an individual is found that passes all of the cases in $T_A$, we test the individual on all of the cases in $T \setminus T_A$; if it also passes all of them, then it is a training set solution and GP terminates. Otherwise, we select a random case in $T$ that the individual does not pass, add it to $T_A$, and continue evolution. Note that if multiple individuals in a generation pass all of the cases in $T_A$, each of them goes through this process, potentially adding multiple new cases to $T_A$ for the next generation.

Given that we already have a set of training cases, why does iCDGP use a smaller, likely less-informative set of active training cases?
As with other approaches based on the sub-sampling of training cases (such as down-sampled lexicase and cohort lexicase selections~\cite{Hernandez:2019:GECCOcomp,Ofria:2019:GPTP}), a smaller active training set allows iCDGP to perform fewer program executions per generation, making each generation computationally cheaper than if using the full set of training cases. In our experiments, we compare methods based on the same maximum number of program executions, allowing iCDGP to run for more generations than standard GP while using the same total program executions. Additionally, the CDGP idea of adding a counterexample case to $T_A$ that the best individual does not pass allows it to augment the training set in ways that specifically direct GP to solve difficult parts of the problem that have not yet been solved by the evolving population.

In our experiments, we test several variants of iCDGP to try to improve it. One such variant uses a fitness threshold $q$ to determine when to add a new counterexample, a variant introduced in formal CDGP~\cite{Bladek:EC}.
For iCDGP, this triggers the system to test the individual on all of $T$, and then add a case to $T_A$ that the individual does not pass. It is possible that the individual passes all of the cases in $T$ that are not already in $T_A$; in this case, $T_A$ does not change.



We designed another variant of iCDGP to address an issue that we discovered when analyzing our results, as presented in Section~\ref{sec:results}. In particular, many times iCDGP cannot find a program that passes all (or even a sufficiently high percentage to exceed a fitness threshold) of the active training set without passing all training cases. It may still be beneficial to add new training cases to $T_A$ to provide GP with more information to guide search. Thus we created a variant of iCDGP, called generation-based case additions, that
adds a new training case to $T_A$ every $d$ generations after the last case was added (whether through this process or an individual passing all of $T_A$). In order to select a case that provides better information for the search, we evaluate the best individual in the population (i.e. the one that passes the most cases in $T_A$, with ties broken at random) on all cases in $T$, and choose a random case that it does not pass.



Finally, we test a variant that sets a maximum number of cases to add to $T_A$. This variant prevents $T_A$ from getting too large, which may be undesirable since it reduces the number of generations that GP can evaluate before using up the program execution budget. When adding a case that would increase the size of $T_A$ past the given limit, we first remove the case in $T_A$ that is passed by the most individuals in the population. In this way, we can remove cases that provide less useful direction to search while adding cases not passed by the best individuals.

\section{Experiment Design}
\label{sec:methods}

In this study we focus on general program synthesis problems, which require the GP system to generate programs that have similar qualities to the types of programs we expect humans to write. For our experiments we use 12 problems with a range of difficulties selected from the PSB1 benchmark suite~\cite{Helmuth:2015:GECCO}. These problems use different data types as inputs and outputs, and many require iteration or recursion and conditional execution to solve.

Each program synthesis problem is defined by a training set $T$ of input/output examples, as well as an unseen test set that is used to test for generalization of solutions to unseen data.\footnote{The datasets for these problems are available at \url{https://github.com/thelmuth/program-synthesis-benchmark-datasets}.}
As discussed above, when a program is found that passes all of the cases in the active training set $T_A$, it is tested on all cases in $T$; if it passes those as well, GP terminates. We then automatically simplify the program using a process that shrinks program sizes without changing its behavior on $T$; this simplification has been shown to increase generalization on the benchmark problems used in this study~\cite{Helmuth:2017:GECCO}. The simplified program then undergoes generalization testing on the test set; if it passes all of the unseen test cases, we consider it a successful run. If a program does not pass the test set, or if a run terminates from reaching the execution limit, it is marked a failure. We use a chi-square test with a 0.05 significance level to test for significant differences in success rates.

Our experiments use PushGP, which evolves programs in the Push programming language~\cite{1068292}. Push, designed specifically for use in GP, uses a set of typed stacks to store data manipulated by a program. Push programs are hierarchical lists containing data literals, which are pushed onto stacks when encountered in programs, and instructions, which take their inputs from specifics stacks and return their results to the stacks. We use an implementation of PushGP written in Clojure in our experiments.\footnote{\url{https://github.com/lspector/Clojush}}

\begin{table}[t]
    \centering
    \caption{PushGP system parameters.}
    \label{table:parameters}
    \begin{tabular}{l r}
        \toprule
        \textbf{Parameter} & \textbf{Value} \tabularnewline
        \midrule
        population size & 1000 \tabularnewline
        max generations for runs using full training set & 300 \tabularnewline
        parent selection & lexicase \tabularnewline
        genetic operator & UMAD \tabularnewline
        UMAD addition rate & 0.09 \tabularnewline
        initial size of $T_A$ & 10 \tabularnewline
        \bottomrule
    \end{tabular}
\end{table}

The PushGP system parameters used in our experiments are given in Table~\ref{table:parameters}. Individual genomes are stored in the Plushy representation, and translated into Push programs for execution. We use uniform mutation with additions and deletions (UMAD) as our only genetic operator, making all children through mutation only, since this mutation has produced the best known results when using PushGP on these benchmark problems~\cite{Helmuth:2018:GECCO}. We use the size-neutral version of UMAD, adding new instructions before or after 9\% of instructions in the parent. Additionally, we use lexicase selection for parent selection~\cite{Helmuth:2015:GECCO,Helmuth:2015:ieeeTEC}.

Lexicase selection has one peculiar characteristic with respect to iCDGP: when an individual is found that passes all cases in $T_A$ but not all those in $T$, we add a new case to $T_A$. However, the selection of parents for the next generation is based on $T_A$ before the new case is added, since that is the set of cases that the population is evaluated on. In lexicase selection, if an individual passes all cases considered for selection and no other individual does, then it will be selected in every single parent selection event that generation. Thus we might expect substantial drops in population diversity each time we add a new case to $T_A$. We investigate the implications of this interaction between lexicase selection and iCDGP empirically in Section~\ref{sec:effects-on-diversity}.

\begin{table}[t]
    \centering
    \caption{Full training set size and program execution limit for each problem.}
    \label{table:execution-limits}
    \begin{tabular}{p{7cm} r r}
        \toprule
        \textbf{Problems} & \textbf{Training Size} & \textbf{Executions} \tabularnewline
        \midrule
        Compare String Lengths, Double Letters, Mirror Image, Replace Space With Newline, Smallest, String Lengths Backwards, Syllables & 100 & 30,000,000 \tabularnewline
        \midrule
        Last Index of Zero, X-Word Lines & 150 & 45,000,000 \tabularnewline
        \midrule
        Negative to Zero, Scrabble Score & 200 & 60,000,000 \tabularnewline
        \midrule
        Vector Average & 250 & 75,000,000 \tabularnewline
        \bottomrule
    \end{tabular}
\end{table}


Since iCDGP executes fewer programs per generation, all of our PushGP runs are limited by the number of program executions they allow, equivalent to using a population size of 1000 and 300 maximum generations for a full training set. This ensures that all methods receive the same number amount of computation. Since iCDGP uses fewer training cases to evaluate each individual, it runs for more generations to make up the same number of program executions. The number of training cases in the full training set varies per problem, so the maximum program execution limits also vary per problem, and both are given in Table~\ref{table:execution-limits}.

\section{Results} \label{sec:results}

\begin{table}[t]
\centering
\rowcolors{3}{gray!15}{white}
\begin{tabular}{l | rrrrrr}
\toprule
\textbf{Problem }  & \textbf{ Full} & \textbf{ iCDGP} & \textbf{ q = 0.8} & \textbf{ d = 25} & \textbf{ d = 50} & \textbf{ d = 100}\\
\midrule
CSL              & 32  & \sigWorse{13} & 41 & 30 & 20 & \sigWorse{18}  \\
DL               & 19  & 24 & 26 & \textbf{37} & 32 & 29  \\
LIoZ             & 62  & \sigWorse{41} & 56 & 63 & 62 & 65  \\
MI               & 100 & 98 & \sigWorse{89} & \sigWorse{93} & 98 & 96  \\
NTZ              & 80  & 76 & 80 & 79 & 80 & 81  \\
RSWN             & 87  & 94 & \textbf{96} & 91 & \textbf{96} & 91  \\
SS               & 13  & 15 & \textbf{30} & \textbf{50} & \textbf{31} & \textbf{42}  \\
Smallest         & 100 & \sigWorse{93} & \sigWorse{93} & 96 & 95 & 95  \\
SLB              & 94  & 90 & 90 & \sigWorse{83} & 87 & 87  \\
Syl              & 38  & \sigWorse{24} & 44 & \textbf{69} & \textbf{62} & 49  \\
VA               & 88  & 87 & 89 & \textbf{97} & \textbf{97} & \textbf{98}  \\
XWL              & 61  & \textbf{75} & \textbf{82} & \textbf{85} & \textbf{89} & \textbf{87}  \\
\bottomrule
\end{tabular}
\caption{The number of successes out of 100 GP runs. All results are compared to \textbf{Full}; results that are significantly better are in \textbf{bold}, and results that are significantly worse are \underline{\textit{underlined and italicized.}} \textbf{Full} is GP using the full training set. \textbf{iCDGP} is the standard version of iCDGP. \textbf{q = 0.8} is the variant of iCDGP that adds a case any time an individual passes more cases than the fitness threshold of $q = 0.8$. The three columns labeled with values for \textbf{d} are the variant that adds a case to the training set every $d$ generations.}
\label{table:compareToFull}
\end{table}



\begin{figure}[t]
    \centering
    \includegraphics[width=0.8\textwidth]{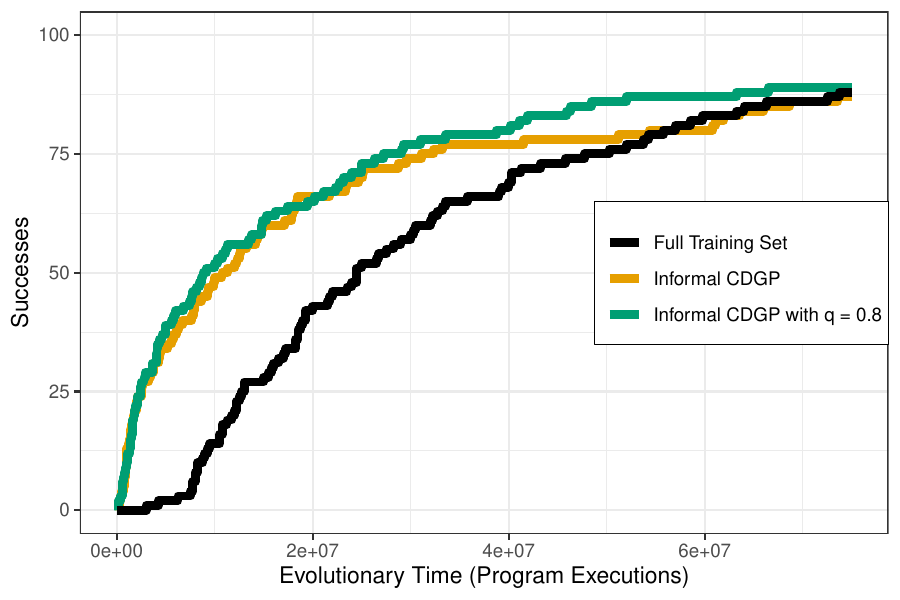}
    \caption{Cumulative number of successful GP runs on the Vector Average problem over evolutionary time, as measured by program executions.}
    \label{figure:va-cdgp}
\end{figure}

We first present results comparing iCDGP to GP using the full training set. The first three columns of Table~\ref{table:compareToFull} give the number of successful GP runs out of 100 using a full training set, iCDGP, and iCDGP with a fitness threshold of $q = 0.8$. First, note that iCDGP performed a bit worse than using the full training set, including significantly worse on four problems while only significantly better on one. On the other hand, iCDGP using a fitness threshold performed significantly better than the full training set on three problems while only performing significantly worse on two, showing the benefits of adding cases before finding a solution on the training set.

Despite producing no notable improvement in performance on these benchmark problems over using the full training set, we did notice that the solutions that iCDGP found often occurred earlier in evolutionary time than with the full training set. For example, Figure~\ref{figure:va-cdgp} shows the cumulative number of successes over evolutionary time on the Vector Average problem; this plot is representative of many of the other problems we observed. Both of the iCDGP methods (with and without a fitness threshold) find many solutions quite early in their runs, reaching 50 successes around 10 million program executions, at which point the full training set has produced only 14 solutions. However, the full training set catches up over evolutionary time, reaching about the same number of successes by the time it hits the maximum number of program executions.
iCDGP's ability to find solutions earlier likely stems from it executing many fewer programs per generation, allowing it to produce more generations (and therefore explore more programs) within the same number of program executions.
The rapid production of solutions provides one argument for using iCDGP.

\subsection{Variant: Generation-Based Case Additions}
\label{sec:results:generation-based-case-additions}

With generation-based case additions, we add a new case to $T_A$ every time $d$ generations have passed without a new case being added otherwise. We tested three settings for $d$: 25, 50, and 100 generations. Note that failed iCDGP runs often finished after 1000 to 3000 generations, depending on how many cases are added to $T_A$.


The last three columns of Table~\ref{table:compareToFull} present the number of successful runs for different settings of $d$. iCDGP with generation-based additions performed similarly to or better than iCDGP without them on every problem for all three settings of $d$. While all three performed sometimes better and sometimes worse than each other, we will concentrate on $d = 50$ here, which was significantly better than the full training set on five problems while never performing significantly worse.

Generation-based case additions turns iCDGP's questionable benefits into clear ones. They also present an improvement over using a fitness threshold, likely because they allow for the addition of new cases without an individual having to reach the threshold. 
Future work would be needed to determine whether selecting a new case that is not passed by the best individual helps, or if adding any random case from $T \setminus T_A$ would be sufficient. 





\subsection{Variant: Maximum size of active training set}


\begin{table}[t]
\centering
\rowcolors{3}{gray!15}{white}
\begin{tabular}{l | rrr | rr}
\toprule
\textbf{Problem }  & \textbf{ d = 50} & \textbf{ Max of 10} & \textbf{ Max of 20 } & \textbf{ Static} & \textbf{ DSL}\\
\midrule
CSL              & 20 & 21                & \textbf{53}    & \sigWorse{0}             & 25  \\
DL               & 32 & 46                & 37             & \sigWorse{4}             & \textbf{72}  \\
LIoZ             & 62 & 66                & 58             & \sigWorse{7}             & 68  \\
MI               & 98 & 98                & 97             & \sigWorse{13}            & 99  \\
NTZ              & 80 & 77                & 80             & \sigWorse{31}            & 84  \\
RSWN             & 96 & 88                & 95             & \sigWorse{57}            & 96  \\
SS               & 31 & \sigWorse{2}      & \sigWorse{15}  & \sigWorse{13}            & \sigWorse{18}  \\
Smallest         & 95 & 97                & 94             & \sigWorse{40}            & 99  \\
SLB              & 87 & 93                & 85             & \sigWorse{35}            & \textbf{96}  \\
Syl              & 62 & \sigWorse{36}     & 52             & \sigWorse{9}             & 61  \\
VA               & 97 & 95                & 95             & \sigWorse{71}            & 100 \\
XWL              & 89 & 91                & 94             & \sigWorse{35}            & 95  \\
\bottomrule
\end{tabular}
\caption{The number of successes out of 100 GP runs. All results are compared to \textbf{d=50}; results that are significantly better are in \textbf{bold}, and results that are significantly worse are \sigWorse{underlined and italicized.} \textbf{Max of 10} and \textbf{Max of 20} are the variant of iCDGP that caps the size of $T_A$ at 10 or 20 respectively. \textbf{Static} uses a fixed training set, for the experiment in Section~\ref{subsec:static}. \textbf{DSL} is down-sampled lexicase selection, as discussed in Section~\ref{subsec:compare-with-dsl}.}
\label{table:compareToD}
\end{table}

Since it appears that some of the benefits of iCDGP derive from the fact that it uses a small number of cases per program evaluation,
we to hypothesize that its performance might improve if the number of cases were capped.
For the experiments that produced the results shown in the columns \textbf{Max of 10} and \textbf{Max of 20} in Table~\ref{table:compareToD}, we began with the version of iCDGP using generation-based case additions every 50 generations.
To this configuration we added a mechanism that removes a case each time a new case is added, once the number of cases has reached a pre-specified maximum.
Specifically, whenever we add a case that would increase the size of $T_A$ past the given limit, we first remove the case in $T_A$ that is passed by the most individuals in the current population, with the intention to remove cases that provide less useful direction to search.


As can be seen in Table~\ref{table:compareToD}, limiting $T_A$ to 10 cases degrades problem-solving significantly on the Scrabble Score and Syllables problems. Limiting $T_A$ to 20 cases produces significantly worse results on Scrabble Score, but significantly better on Compare String Lengths. Neither of these results suggests that limiting the size of $T_A$ deserves recommendation.

\subsection{Benefits of Counterexample Cases} \label{subsec:static}

One may wonder whether the benefits we have demonstrated with iCDGP come entirely from having a small active training set $T_A$ on which we evaluate each individual, reducing the number of program executions per generation. In other words, it is possible that adding counterexample cases to $T_A$ provides no benefits. To test this hypothesis, we conducted a set of runs that use a static active training set consisting of 10 random cases, the same number as the size of $T_A$ at the start of our iCDGP runs. 
Note that the only functional difference in these methods happens when a program is found that passes all cases in $T_A$. 
When a run with a static training set finds a program that passes all cases in $T_A$, it is simply tested for generalization.


Table~\ref{table:compareToD} compares the number of successes produced by GP with iCDGP adding a case every $d = 50$ generations to GP with a static training set. iCDGP is significantly better on every problem, often by huge margins. These differences highlight the importance of iCDGP's additions of counterexample cases to $T_A$.

\subsection{Effects on Population Diversity} 
\label{sec:effects-on-diversity}

\begin{figure*}[t]
    \centering
    \includegraphics[width=0.9\textwidth]{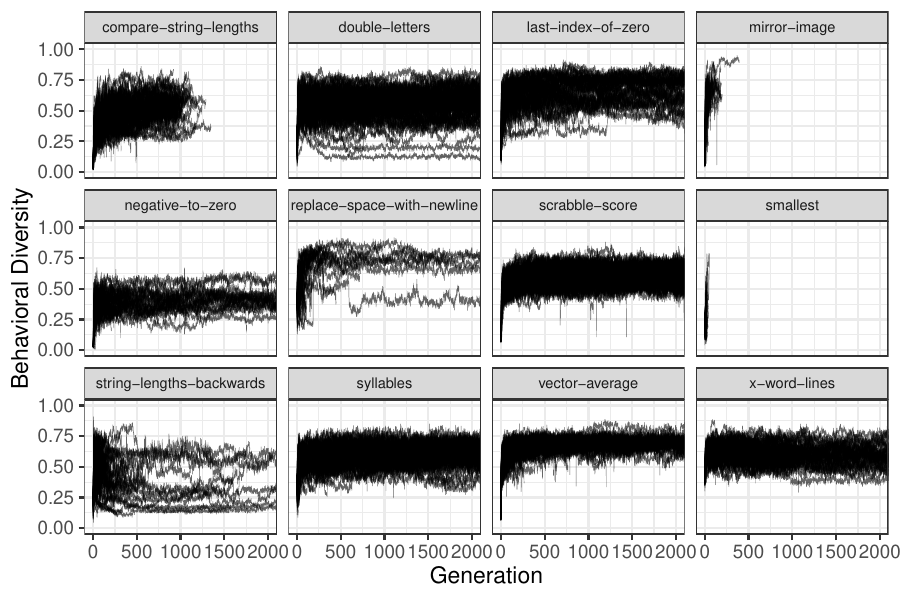}
    \caption{Population behavioral diversity for standard iCDGP runs, cropped at 2000 generations. Each run is plotted separately. Note that all runs for Mirror Image and Smallest found solutions early in the runs, and Compare String Lengths runs ended earlier than others because they often added many training cases to $T_A$.}
    \label{figure:diversity-iCDGP}
\end{figure*}

We are interested in the effects of iCDGP, especially with lexicase selection, on population diversity. In particular, as we discussed in Section~\ref{sec:methods}, when an individual passes all cases in $T_A$ and iCDGP adds another case, lexicase selects that individual as the parent of every child in the next generation. If more than one such individual is found in the same generation, then the selections will be randomly distributed among them.

These \textit{hyperselection} events will mean that every individual created after such an event will be a descendant of the hyperselected individual. Previous work studied hyperselection events with lexicase selection, but specifically individuals that received 5-10\% of the selections in a generation, not 100\% of them~\cite{Helmuth:2016:GECCO}. This work found that hyperselection events had no noticeable effects on problem-solving performance, and only caused brief reductions in population diversity. Another study of lexicase selection found that it is able to quickly recover population diversity in situations when the population had low diversity~\cite{Helmuth:2016:GECCOcomp}. Here we examine whether these hyperselection events have detrimental effects on population diversity when using iCDGP.

We measure population diversity in terms of \textit{behavioral diversity}, or the proportion of distinct behavior vectors produced by a population~\cite{JacksonPaper}; a \textit{behavior vector} is a list of outputs that the program produces when run on the cases in $T_A$. Figure~\ref{figure:diversity-iCDGP} plots the behavioral diversity of every single iCDGP run on all 12 problems as a separate line. Looking closely, there are clearly instances where population diversity drops drastically in one generation\footnote{Note that the diversity does not go all the way to 0, since even if one parent created all of the children in the next generation, some of those children are likely to display different behaviors from the parent.}, with many in the Scrabble Score and Vector Average problems, and a few in most of the other problems. Most of these drops in diversity follow one of two patterns: a solution to the full training set $T$ is found in the next generation, leading to a line that drops down and then ends; or a quick increase in diversity over a few generations back to levels seen before the drop. On the other hand, we see little evidence for sudden drops in diversity leading to extended stretches of low diversity. 

So, while these hyperselection phenomena do occur when an individual passes all cases in $T_A$, there does not seem to be corresponding long-term detrimental effects on population diversity. Lexicase selection may be the cause and the cure, as its case-by-case effects provide boosts in diversity following hyperselection.

\subsection{Number of Active Cases}

\begin{figure}[t]
    \centering
    \includegraphics[width=0.9\textwidth]{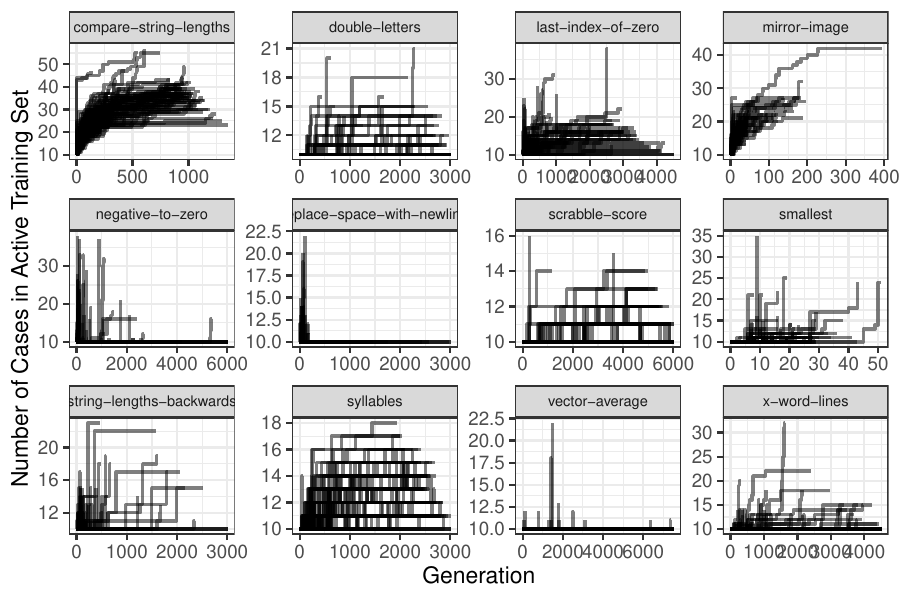}
    \caption{The number of training cases in the active training set $T_A$ for iCDGP runs. Each run is plotted separately. Note that no cases are ever removed, so each line can only increase. Also note different x-axis and y-axis scales per problem.}
    \label{figure:active-cases-iCDGP}
\end{figure}

\begin{figure}[t]
    \centering
    \includegraphics[width=0.9\textwidth]{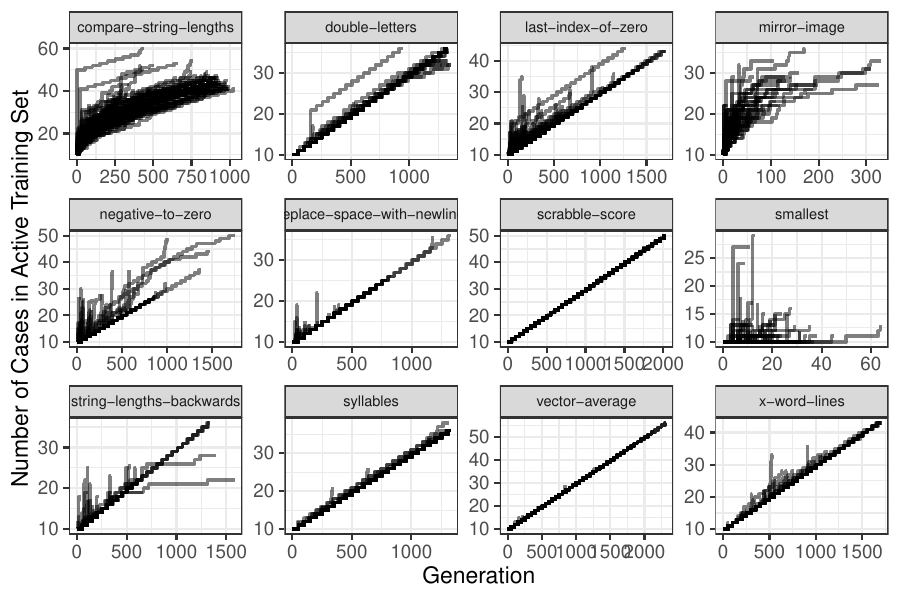}
    \caption{The number of training cases in the active training set $T_A$ for iCDGP with $d = 50$, adding a new case every 50 generations. Each run is plotted separately. Note that no cases are ever removed, so each line can only increase. Also note different x-axis and y-axis scales per problem.}
    \label{figure:active-cases-add50}
\end{figure}

In order to get a better idea of how often iCDGP adds cases to $T_A$, we plot the number of cases in $T_A$ over evolutionary time for standard iCDGP in Figure~\ref{figure:active-cases-iCDGP} and for the version that adds a case every 50 generations in Figure~\ref{figure:active-cases-add50}. 

For iCDGP, we see many different patterns of when and how many cases are added to $T_A$. For example, Compare String Lengths and Mirror Image are the only problems in our benchmark set with Boolean-valued outputs, making them easier to pass all cases in $T_A$ without passing all of $T$ than problems with outputs coming from a wider domain, such as numbers or strings. Thus we find it unsurprising that these two problems consistently see the largest growth in $T_A$. Other problems add cases at different rates, with Replace Space with Newline, Vector Average, and Negative to Zero falling at the other extreme, where a few runs added quite a few cases early and were solved, while the rest never added any cases.

The stair-step pattern of sizes of $T_A$ in Figure~\ref{figure:active-cases-add50} reflects the cases that are added to $T_A$ after 50 generations since a case was lasted added. Some problems, corresponding roughly with the problems in Figure~\ref{figure:active-cases-iCDGP} that add few cases, rarely if ever add a case besides every 50 generations. Other problems still seem to add quite a few cases for individuals that pass all of $T_A$. We find no correlation between these two types of problems and those at which this version of iCDGP performs better compared to the standard. The performance improvement seen when adding a case every 50 generations seems to benefit both kinds of problems.

\subsection{Comparison with down-sampled lexicase Selection} \label{subsec:compare-with-dsl}

We compare iCDGP to down-sampled lexicase selection, using results from~\cite{Helmuth2020explaining}. To ensure fairness of the comparison, we only consider down-sampled lexicase with down-sample rates which result in 10 cases being evaluated each generation, which is the size of iCDGP's active set $T_A$. For instances where no such down-sampling rate existed (Last Index of Zero, Vector Average, and X-Word Lines), we used the results from a down-sampling rate that resulted in just a few more than 10 cases being evaluated per generation. 

We found that iCDGP (using $d = 50$) is competitive to down-sampled lexicase selection, with a comparison in Table~\ref{table:compareToD}. Of the 12 test problems, iCDGP performed significantly better on one, while significantly worse on only two. The results from down-sampled lexicase selection are among the best results achieved on these PSB1 problems, giving iCDGP a strong comparison to the state-of-the-art.

\section{Conclusions and future work}

We conclude that informal counterexample-driven genetic programming (iCDGP) advances the state of the art for software synthesis by GP. It builds on the recent advance provided by formal CDGP, but it is likely to be more widely applicable because it does not require a formal specification of solutions to the target problem. The same set of test inputs that would be used for traditional GP can be used for iCDGP, with the only difference being how they are used. Specifically, iCDGP begins with a small initial subset of the cases, and augments the subset with counterexamples whenever an individual passes all of the current cases.
We introduce new variants of iCDGP that experimentally outperform the standard version. We recommend using the version that adds a new case to the active set $T_A$ every $d$ generations, ensuring that cases are added even if no program is found that passes all cases in $T_A$. This variant performed best for iCDGP, and future work could investigate its use in CDGP with formal constraints.

Although we explored several variants of iCDGP, we anticipate other variants to emerge which may outperform ones presented here.
Future work should focus on conducting further analyses of the underlying evolutionary dynamics that are responsible for the success of the technique to guide us in developing improvements. We have presented here some preliminary data on behavioral diversity and numbers of cases in $T_A$ over evolutionary time, but many other aspects of these runs can be investigated, and other variants of the technique tested to explore hypotheses about the reasons that it works.
For example, with respect to generation-based additions, it would be useful to learn wither it is important to include new cases that are not passed by the best individual, or if the same benefit would result, more simply, from adding any random case from $T \setminus T_A$.

\section*{Acknowledgements}
We thank the members of the PUSH lab for discussions that improved this work.
This material is based upon work supported by the National Science Foundation under Grant No. 2117377. Any opinions, findings, and conclusions or recommendations expressed in this publication are those of the authors and do not necessarily reflect the views of the National Science Foundation.

%
%
%
\bibliographystyle{splncs04}
\bibliography{icdgp-bib}

\end{document}